\documentclass[10pt,twocolumn,letterpaper]{article}

\usepackage[pagenumbers]{cvpr} 

\usepackage{graphicx}
\usepackage{amsmath}
\usepackage{amssymb}
\usepackage{booktabs}
\usepackage[accsupp]{axessibility}  

\usepackage{colortbl}
\usepackage{comment}
\usepackage{bm}
\usepackage{algorithm, algpseudocode}
\usepackage{mathtools}
\usepackage{arydshln}
\usepackage{multirow}

\usepackage{capt-of}

\usepackage[table,xcdraw,dvipsnames]{xcolor}
\usepackage{threeparttable}
\usepackage{url}

\usepackage{a_figures}
\usepackage{b_tables}

\newcommand{\LN}[1]{{\rm LN}(#1)}
\newcommand{\MLP}[1]{{\rm MLP}(#1)}

\usepackage[pagebackref,breaklinks,colorlinks]{hyperref}

\usepackage[capitalize]{cleveref}
\crefname{section}{Sec.}{Secs.}
\Crefname{section}{Section}{Sections}
\Crefname{table}{Table}{Tables}
\crefname{table}{Tab.}{Tabs.}

\def\cvprPaperID{7143} 
\def\confName{CVPR}
\def\confYear{2023}

\begin{document}

\title{Hierarchical Neural Memory Network for Low Latency Event Processing}

\author{
Ryuhei Hamaguchi$^{1}$,
Yasutaka Furukawa$^{2}$,
Masaki Onishi$^{1}$,
and Ken Sakurada$^{1}$ \\
$^{1}$ National Institute of Advanced Industrial Science and Technology (AIST)\\
$^{2}$ Simon Fraser University \\
{\tt\small ryuhei.hamaguchi@aist.go.jp, furukawa@sfu.ca, onishi@ni.aist.go.jp, k.sakurada@aist.go.jp}
}
\maketitle

\begin{abstract}
This paper proposes a low latency neural network architecture for event-based dense prediction tasks.
Conventional architectures encode entire scene contents at a fixed rate regardless of their temporal characteristics.
Instead, the proposed network encodes contents at a proper temporal scale depending on its movement speed.
We achieve this by constructing temporal hierarchy using stacked latent memories that operate at different rates.
Given low latency event steams, the multi-level memories gradually extract dynamic to static scene contents by propagating information from the fast to the slow memory modules.
The architecture not only reduces the redundancy of conventional architectures but also exploits long-term dependencies.
Furthermore, an attention-based event representation efficiently encodes sparse event streams into the memory cells.
We conduct extensive evaluations on three event-based dense prediction tasks, where the proposed approach outperforms the existing methods on accuracy and latency, while demonstrating effective event and image fusion capabilities.
The code is available at \url{https://hamarh.github.io/hmnet/}.
\end{abstract}

\section{Introduction}
\label{sec:introduction}

Latency matters for many vision applications such as autonomous vehicles or UAVs,
directly affecting their safety and reliability measures.
Latency is also crucial for better user experience in time-sensitive vision applications such as augmented reality, where 
the latency of standard RGB cameras is insufficient.
For example, a vehicle traveling 80km/h will move 74cm within a frame of 
a standard 30fps camera.

Event cameras have extremely low latency.
Unlike standard vision cameras that capture the intensity of all pixels at a fixed rate, event cameras asynchronously record intensity changes of individual pixels. This unique principle leads to extremely low latency (microseconds) and high temporal resolution, along with many other advantages such as high dynamic range and low power consumption.
With such attractive characteristics, 
event cameras are becoming popular input devices for many dense prediction tasks such as semantic segmentation \cite{Alonso2019, Sun2022}, object detection \cite{Perot2020, Schaefer2022, Li2022}, depth estimation \cite{Gehrig2021, Nam2022}, and optical flow 
\cite{Zhu2019, Gehrig2021OpticalFlow}.

\figIntro

Despite the emergence of low latency event-cameras,
few works focused on low latency recognition models dedicated to event-based dense prediction tasks.
Instead, most previous works apply standard CNN architectures equipped with the recurrent modules as a backbone \cite{Carrio2020,Perot2020,Sun2022,Gehrig2021}, resulting in the same latency levels as frame-based models.
Another line of research 
applies Spiking Neural Networks (SNNs) for event data. However, SNNs suffer from low accuracy due to the lack of established training principles \cite{Tavanaei2018,Wu2018} or high latency due to long simulation time steps \cite{Yan2021}.
We need backbone architectures that best exploit low latency and high temporal resolution of event data.

To this end, we propose a Hierarchical Neural Memory Network (HMNet) for low latency event processing.
The key idea is to encode the scene contents at a proper temporal scale depending on their speed.
For this purpose, the proposed network builds multi-level latent memory with different operating rates (Fig.~\ref{fig:intro}).
The low-level memories run fast to quickly encode local and dynamic information, whereas high-level memories perform global reasoning on static information at a lower frequency.
This design 
significantly reduces the computational loads in contrast to conventional methods that runs the entire forward path every time.
The paper also proposes 
an Event Sparse Cross Attention (ESCA) that directly injects sparse event streams into dense memory cells with minimal information loss.

We conduct extensive evaluations on 
three event-based vision tasks (object detection, semantic segmentation, and monocular depth estimation) as well as event-image fusion task.
Experimental results show that HMNet 
outperforms existing methods while reducing latency by 40\%-50\%.

\section{Related Works}
\label{sec:related_works}
Event-based dense prediction often consists of the following two steps:
1) A construction of frame-based representations for events and 2) feature extraction using a backbone CNN followed by a task-specific head. The section reviews related works in these two steps, followed by discussion on specialized methods without these steps.

\vspace{0.1cm}
\noindent
\textbf{Event representation.}
An efficient and informative neural event representation
is the key to effective event processing.
Due to its sparse and asynchronous nature, event data are not directly applicable to many DNN models with dense-grid representations.
The simplest solution is to convert events into histogram images \cite{Maqueda2018, Nguyen2019}, but the method completely discards rich temporal information of events.
In order to maximize the information, many event representations are proposed. 
Timestamp images \cite{Park2016} and Time Surface \cite{Lagorce2017, Chen2020} encode events based on the most recent timestamp at each pixel.
HATS \cite{Sironi2018} further improved the methods by additionally encoding timestamps of past events.
Voxel Grid \cite{Zhu2019} constructs spatiotemporal voxel representation by discretizing the time domain using a bilinear kernel.

Recently, more data-adaptive methods have been proposed \cite{Gehrig2019, Cannici2020, Liu2022, Li2022}.
EST \cite{Gehrig2019} generalized the above methods as Event Spike Tensor and further makes the kernel trainable end-to-end from data. Matrix-LSTM \cite{Cannici2020} applies pixel-wise LSTM to extract temporal features from the event stream. AED \cite{Liu2022} and ASTMNet \cite{Li2022} improve Voxel Grid by adaptively adjusting kernel size and sampling position according to the event density.

In contrast,
the proposed method is 
1) Data-adaptive: the attention adaptively picks up necessary information from events based on the current memory state,
2) Robust to noise: the attention can filter out noisy events, and
3) High frame rate: the memory-based representation can work at a short interval (\eg 5ms), contrary to the previous methods that require a long time (\eg 50ms) to accumulate enough data (excluding Time Surface and Matrix-LSTM).

\vspace{0.1cm}
\noindent
\textbf{Backbone architecture.}
A CNN architecture with a recurrent module is a popular architecture for event-based dense prediction tasks.
Perot \etal \cite{Perot2020} built a CNN architecture with stacked ConvLSTM \cite{Shi2015} for object detection. Li \etal \cite{Li2022} also combined lite-weight LSTM with VGG16 \cite{Simonyan2015}.
The recurrent architectures based on U-Net \cite{Ronneberger2015} or stacked hourglass network \cite{Newell2016} are also used for monocular depth estimation \cite{Carrio2020, Gehrig2021} and semantic segmentation \cite{Sun2022}.

Several works proposed a backbone architecture for asynchronous event processing to better exploit the sparse and asynchronous nature of the event data.
Messikommer \etal \cite{Messikommer2020} proposed a sparse convolutional network for asynchronous event processing.
Schaefer \etal \cite{Schaefer2022} applied a graph convolutional neural network for event processing and proposed the efficient and asynchronous graph update rule for events.
While these methods can dramatically reduce computational complexity, the performance is insufficient compared to conventional synchronous models.

Exploiting the sparsity of event data has been popular for classification tasks.
EventNet \cite{Sekikawa2019} treated event data as a point cloud and applied MLPs event-by-event like PointNet \cite{Qi2017}. By converting the trained MLPs into a lookup table, the network runs extremely fast at inference time. Event Transformer \cite{Li2022evtrans} also takes raw events as inputs and applies specialized transformer layers to extract spatiotemporal features for a group of events. Although these methods perform well on the classification task, applying them to dense prediction tasks is difficult since making dense predictions from the event-wise sparse features is not trivial.

Sabater \etal \cite{Sabater2022} proposed to update an internal memory using sparse patch-based representation extracted from the event Voxel Grid.
This work is the most related to ours, but the method still relies on the conventional Voxel Grid representation, which is sub-optimal. Our method directly writes raw events into memory through cross-attention, which is more data-adaptive.
Hierarchical Temporal Memory (HTM) \cite{Hawkins2011} is conceptually related to our work.
HTM consists of Spatial Pooling and Temporal Pooling that are conceptually similar to convolution and recurrent layers.
HTM is more similar to RNNs and it operates at a single rate. In contrast, the proposed model operates at multiple rates, which is a key for low-latency processing.

\section{Hierarchical Neural Memory Network}
\label{sec:method}

\figHMNet
\figMemory
\figTimingChart

Our idea for low latency event processing is a multi-rate network architecture.
A scene often contains objects with varying motion speeds.
A network should run fast for high-speed motions, but conduct careful or global reasoning for slowly moving objects or scene context analysis.

To achieve this, the proposed network builds a temporal hierarchy using multi-level latent memories $\{\bm{z}_1,...,\bm{z}_L\}$ that operate in parallel at different rates.
Fig.~\ref{fig:hmnet} shows an overview of the proposed network.
The memories are stacked such that their operating rate decreases from $\bm{z}_1$ to $\bm{z}_L$.
$\bm{z}_1$ writes incoming events into its state (event-write) and quickly extracts local and dynamic information with a shallow network (update). The features are then propagated to higher memories (up-write) where global and static information is extracted with deeper networks (update).
The network also has a top-down path (down-write) that enables low-level memories to exploit the contextual information to recognize dynamic motion accurately.

At the end of the operating cycle, each memory computes output features (readout) and puts them into a latent buffer. At every time step, the task head computes predictions from the features inside the latent buffer,
exploiting low latency information and global context simultaneously.

Below, Sec.~\ref{subsec:latent_memory} explains the basic operations of latent memories. Sec.~\ref{subsec:event_write} proposes a method for writing raw event data into the latent memory. Sec.~\ref{subsec:multi-level_latent_memory} explains the overall operation flow and timing of the proposed network. Finally, Sec.~\ref{subsec:sensor_fusion} introduces the extension of the proposed network for multi-sensory inputs with varying frame rates.

\subsection{Latent memory}
\label{subsec:latent_memory}
We have a hierarchy of latent memory where memory cells update memories by attention mechanisms.
Each memory has four operations: ``{\it up-write}'', ``{\it down-write}'', ``{\it update}'', and ``{\it readout}''. 
The memory $\bm{z}_1$ has an additional operation, ``{\it event-write}'', to write raw event data into the memory state.
The architectures for each operation are depicted in Fig.~\ref{fig:memory}. Sec.~\ref{subsec:event_write} explains event-write in detail.
Below, we explain the other four operations.

\vspace{0.1cm}
\noindent
\textbf{Up-write.}
The operation writes the memory state of the previous level $\bm{z}_{l-1}$ into current state $\bm{z}_l$:
\begin{eqnarray}
\bm{z}_l \leftarrow F_{w}^{\uparrow}(\bm{z}_l, \bm{z}_{l-1}).
\end{eqnarray}
To aggregate features from $\bm{z}_{l-1}$, the write function $F_{w}^{\uparrow}$ applies the window based multi-head cross-attention (W-MCA) that is our extension of W-MSA \cite{Liu2021} for cross-attention:
\begin{eqnarray}
\label{eq:upwrite_attn}
\hat{\bm{z}_l} &=& \text{W-MCA}(\bm{z}_l, G_{down}(\bm{z}_{l-1})) + \bm{z}_l, \\
\label{eq:upwrite_mlp}
\bm{z}_l &\leftarrow& \MLP{\LN{\hat{\bm{z}_l}}} + \hat{\bm{z}_l},
\end{eqnarray}
where LN is a Layer Normalization \cite{Ba2016}.
MLP consists of two conv $1\times 1$ layers with a GELU activation function \cite{Hendrycks2016} in between.
$G_{down}$ is a strided convolutional block for downsampling.

\vspace{0.1cm}
\noindent
\textbf{Down-write.}
The operation writes the memory state of the next level $\bm{z}_{l+1}$ into current state $\bm{z}_l$:
\begin{eqnarray}
\bm{z}_l \leftarrow F_{w}^{\downarrow}(\bm{z}_l, \bm{z}_{l+1}).
\end{eqnarray}
$F_{w}^{\downarrow}$ is almost the same as Eq.~\ref{eq:upwrite_attn},~\ref{eq:upwrite_mlp} except the down/up-sampling operation. $F_{w}^{\downarrow}$ applies $G_{down}$ on $\bm{z}_l$ and upsamples the output of W-MCA with a bilinear upsampling layer (see Fig.~\ref{fig:memory}b,~\ref{fig:memory}c).

\vspace{0.1cm}
\noindent
\textbf{Update}.
The operation updates the internal state of the memories:
\begin{eqnarray}
\bm{z}_l \leftarrow F_u(\LN{\bm{z}_l})
\end{eqnarray}
The function $F_u$ consists of several residual layers \cite{He2016}.
For high-level memories, a large number of layers can be used to encode high-level semantic information. On the other hand, for low-level memories, a few layers are used to prioritize latency.
Specifically, we set the number of residual layers as $\{1, 3, 9\}$ for the memories $\{\bm{z}_1, \bm{z}_2, \bm{z}_3\}$.

\vspace{0.1cm}
\noindent
\textbf{Readout}.
The operation extracts an output feature $\bm{o}_l$ from the internal state of $\bm{z}_l$:
\begin{eqnarray}
\bm{o}_l = F_{ro}(\LN{\bm{z}_l})
\end{eqnarray}
The function $F_{ro}$ consists of conv $1\times 1$ followed by a Group Normalization \cite{Wu2018GroupNorm} and a SiLU activation function \cite{Elfwing2018}.

\subsection{Event write}
\label{subsec:event_write}
Event cameras trigger an event asynchronously at each pixel whenever an accumulated brightness change exceeds a certain threshold.
The output of the event cameras is a stream of asynchronous events $\mathcal{E} = \{\bm{e}_i\}_{i=1}^{N}$, where each event $\bm{e}_i = \{t_i,x_i,y_i,p_i\}$ encodes four values: the timestamp $t_i$, the pixel location $(x_i, y_i)$, and the polarity $p_i\in\{1,-1\}$ that represent the direction of brightness change (increase or decrease).

To embed the sparse event stream into dense memory cells, we propose an Event Sparse Cross Attention (ESCA).
It features high efficiency and robustness to noise by 
exploiting the sparsity of events through window-based attention and introducing a trainable ``event gate'' inside the attention computation.

At every time step $t_n$, the network receives events $\mathcal{E}_n = \{\bm{e}_i\}_{i=1}^{N}$ triggered during $[t_{n-1},t_n]$ as inputs.
Let $\bm{z}_{jk}$ be the memory cell at a spatial location $(j,k)$ and $s$ be the global stride of the memory cells; we define a $s\times s$ spatial window $\mathcal{W}_{jk}$ for each memory cell $\bm{z}_{jk}$. ESCA calculates cross attention between the memory cell $\bm{z}_{jk}$ and a subset of events $\mathcal{E}_n^{(j,k)}$ inside the window $\mathcal{W}_{jk}$ (Fig.~\ref{fig:memory}a).

\begin{equation}
\label{eq:esca}
\bm{h}_{jk} = {\rm CrossAttn}(\bm{z}_{jk},\mathcal{E}_n^{(j,k)}),
\end{equation}
The attention calculation is efficient, because Eq.~\ref{eq:esca} is required only for the memory cells with events inside their local window.
Below, we detail the attention calculation.

\vspace{0.1cm}
\noindent
\textbf{Event embedding}.
Each event $\bm{e}_i \in \mathcal{E}_n^{(j,k)}$ is first transformed into an embedding vector $\bm{d}_i$ using MLPs.
\begin{eqnarray}
\label{eq:embed_xytp}
\bm{d}_t = E_t(\hat{t}_i),\hspace{2mm}\bm{d}_{xy} = E_{xy}(\hat{x}_i, \hat{y}_i),\hspace{2mm}\bm{d}_{p} = E_{p}(p_i), \nonumber \\
\bm{d}_{i} = {\rm LN}([\bm{d}_t, \bm{d}_{xy}, \bm{d}_p]),\quad\quad\quad\quad
\end{eqnarray}
where
$E_t, E_{xy}, E_p$ are MLPs with two hidden layers and a Layer Normalization \cite{Ba2016} in between, and $[,]$ is a concatenation operation.
$(\hat{x}_i, \hat{y}_i)$ is a relative position of the event $\bm{e}_i$ in the window $\mathcal{W}_{jk}$, and $\hat{t}_i$ is a relative timestamp in the time period $[t_{n-1},t_n]$.

\vspace{0.1cm}
\noindent
\textbf{Sparse attention with event gate}.
The attention is then computed between the memory cell $\bm{z}_{jk}$ and the embedding vectors $\bm{d} \in \mathcal{R}^{L\times D}$ of all the $L$ events
in $\mathcal{E}_n^{(j,k)}$:
\begin{eqnarray}
\bm{q}_{jk} = \mathcal{Q}(\bm{z}_{jk}),\hspace{2mm} \bm{K} = \mathcal{K}(\bm{d}),\hspace{2mm} \bm{V} = \mathcal{V}(\bm{d}), \nonumber \\
\bm{h}_{jk} = {\rm softmax}(\bm{q}_{jk}\bm{K}^{T}/\sqrt{D})\bm{V}.\quad
\label{eq:attention}
\end{eqnarray}
where $\mathcal{Q},\mathcal{K},\mathcal{V}$ are the MLPs for computing query, key, and value.
Since raw events are sparse and noisy, Eq.~\ref{eq:attention} 
assigns a large weight to the noisy events when there are a few or no true events inside the window $\mathcal{W}_{jk}$. We address the problem by concatenating a trainable parameter $w$, namely ``event gate'' before the softmax function:
\begin{equation}
\bm{a}_{jk} = {\rm softmax}([\bm{q}_{jk}\bm{K}^{T}/\sqrt{D},w]).
\end{equation}
$\bm{a}_{jk}$ is the attention weights with a length of $(L+1)$.
The last element of $\bm{a}_{jk}$ corresponds to the attention for the event gate that acts as a noise filter;
Since noise will have a weak correlation with the current memory state, the softmax function will assign a high weight to the event gate rather than the noisy events. Thus, the event gate prevents the noise events from being written to the memory state.

Finally, the attention is calculated by omitting the last element of $\bm{a}_{jk}$ and multiplying values $\bm{V}$:
\begin{equation}
\bm{h}_{jk} = [\bm{a}_{jk}]_{1:L}\bm{V}.
\end{equation}

\subsection{Multi-level latent memory}
\label{subsec:multi-level_latent_memory}
Fig.~\ref{fig:timing_chart} describes the overall operation flow.
The network 
has three levels of memory states $\{\bm{z}_1, \bm{z}_2, \bm{z}_3\}$ with an operating cycle of $\{1, 3, 9\}$ time-steps.
Each memory runs on a different thread asynchronously except for the timing of feature transfer.

At every time-step, $\bm{z}_1$ writes incoming event data and executes event-write, update, and readout operations during the time-step.
The internal state of $\bm{z}_1$ is then written to $\bm{z}_2$ at every 3 time-steps.
$\bm{z}_2$ executes ``write'' and ``update'' operations during 2 time-steps, and then executes ``readout'' and message generateion at the last time-step.
$\bm{z}_3$ works similarly to $\bm{z}_2$ but has the longest cycle of 9 time-steps. Note that the message generation of ``down-write'' (Fig.~\ref{fig:memory}c) is executed at the higher memory thread, which saves computations at the time-sensitive lower memory thread.

The latent buffers will store output features that are read out from each memory.
At every time-step, the task head utilizes the features inside the latent buffers to make a prediction, exploiting temporally multi-scale features.

\subsection{Sensor fusion}
\label{subsec:sensor_fusion}
Our architecture can naturally be extended to the multi-sensory inputs with varying operating frequencies. 
We achieve this by simply setting the frequency of memory writes to that of the sensor operations.
Specifically, for event-image fusion, the (RGB) image data is first embedded into a feature map using an image encoder. The feature map is then written to the high-level memory $\bm{z}_3$ every 45ms in the same way as the ``up-write'' operation. The event data is similarly written to the low-level memory $\bm{z}_1$ every 5ms.

\section{Experiments}
\label{sec:experiments}
The section evaluates the accuracy and the latency of the HMNet on three tasks: semantic segmentation, object detection, and monocular depth estimation.

Below, Sec.~\ref{sec:basic_setups} explains the basic settings of the experiments, and Sec.~\ref{sec:semantic_segmentation} to Sec.~\ref{sec:depth_estimation} show the results on each task.
Finally, Sec.~\ref{sec:ablation_study} performs the ablation study to verify the effectiveness of each component of the proposed model.
See supplementary materials for more detailed setups.

\figResutSemsegObjdet

\subsection{Basic setups}
\label{sec:basic_setups}

\noindent
\textbf{HMNet variants.}
We build four variants of HMNet, HMNet-B1/L1/B3/L3.
HMNet-B1 is the simplest form with only one latent memory. HMNet-B3 is its multi-level version with three latent levels. HMNet-L1/L3 is a high-capacity version of HMNet-B1/B3 with increased latent dimensions. Specifically, let $D$ be the dimension of each memory and $H$ be the number of the head for cross attention in the ``write'' operations; the configurations are as follows:
\begin{itemize}
\setlength{\parskip}{-1mm}
\setlength{\itemsep}{-3mm}
\item HMNet-B1: $D=128, H=4$ \\
\item HMNet-L1: $D=256, H=8$ \\
\item HMNet-B3: $D=(128,256,256), H=(4,8,8)$ \\
\item HMNet-L3: $D=(256,256,256), H=(8,8,8)$ \\
\end{itemize}
\vspace{-0.5cm}
For the memories $\{\bm{z}_1, \bm{z}_2, \bm{z}_3\}$, we set the global stride of features $s$ as $\{4,8,16\}$
and the operation cycle as $\{1,3,9\}$ time-steps, where the time-step size is set as $5$ms.

\vspace{0.1cm}
\noindent
\textbf{Baselines.}
We built baselines by 1) Using Time Surface to convert raw events into frame-based representations; 2) Taking a popular backbone network (ResNet, ConvNeXt, or Swin Transformer); and 3) Adding a task head to output predictions.
We also built their recurrent version by inserting ConvGRU \cite{Ballas2016} between stages.

\vspace{0.1cm}
\noindent
\textbf{Event-image fusion.}
We extended HMNet-B3/L3 for event-image fusion as explained in Sec.~\ref{subsec:sensor_fusion}.
For the image encoder, the layers until the stage3 of ResNet-18 \cite{He2016} are used.
For comparison, we also extend the baselines by concatenating the most recent image frame with its inputs.

\vspace{0.1cm}
\noindent
\textbf{Training and inference.}
For training, we used Adam \cite{Kingma2015} and AdamW \cite{Loshchilov2019} as an optimizer, depending on the task. We set the batch size as 16 and the initial learning rate as 2.0e-4 with a cosine learning rate decay.
We used random resize, crop, and horizontal flip for data augmentation.
Unless otherwise mentioned, we test all the methods using Tesla V100 GPU and 20-core Intel(R) Xeon(R) Gold 6148 CPU @ 2.4GHz.

\vspace{0.1cm}
\noindent
\textbf{Justification for hyperparameter tuning.}
We manually determined the hyperparameters for all the experiments.
To ensure that they are not overly tuned, we compared the manual tuning result to automatic tuning with Hyperopt \cite{Bergstra2013} on the object detection task, where we observed similar accuracy (44.7 and 44.4 mAP for manual and automatic tuning).

\subsection{Semantic segmentation}
\label{sec:semantic_segmentation}

\noindent
\textbf{Setups.}
The experiments are conducted on DSEC-Semantic dataset \cite{Sun2022}.
The dataset contains street-scene event data with a resolution of $640 \times 480$ pixels.
It also contains RGB frames recorded at 20Hz. Note that the cameras have different viewpoints; thus, the RGB frames are not perfectly aligned with the event data.
The pseudo pixel-wise annotations are available at 20Hz.
We used the task head of UPerNet \cite{Xiao2018} and trained our models for 60k iterations.

\tbSemsegFusion

\figVisAllTasks

\vspace{0.1cm}
\noindent
\textbf{Results.}
Fig.~\ref{fig:result_semseg} shows the results.
HMNet-B3/L3 outperforms the state-of-the-art ESS model \cite{Sun2022} on both accuracy and latency. Note that ESS utilizes additional training data on the RGB domain, while our methods are trained only on the event data.
HMNet-L3 also outperforms the baselines with strong backbones.
The effectiveness of HMNet-L3 is shown in a qualitative sample in Fig.~\ref{fig:vis_all_tasks} (top row).
The comparison between HMNet-B1/L1 and HMNet-B3/L3 shows the performance gain obtained by stacking multiple memories (+3.8\%/+3.2\%).
The latency of HMNet-B3/L3 can be further reduced by parallel computation when multi-GPU is available (depicted by ``mGPU'' in the figure).

\vspace{0.1cm}
\noindent
\textbf{Event-image fusion.}
We trained the fusion models using events and frames from the left RGB camera. Table~\ref{tab:semseg_rgb_fusion} compares the results in the case with and without RGB fusion.
The baselines perform worse with RGB fusion due to the misalignment between events and RGB frames, whereas HMNet shows improved accuracy.
Even with the misaligned inputs, cross-attention can robustly associate features, enabling the model to leverage both modalities.
We also test the models using the images from the right RGB camera at inference time.
To our surprise, the HMNet models keep the accuracy high, showing their robustness to the different sensor setups.

\subsection{Object detection}
\label{sec:object_detection}

\noindent
\textbf{Setups.}
The experiments are conducted on GEN1 dataset \cite{Perot2020}.
The dataset includes about 40 hours of event data with a resolution of $304\times 240$ pixels. The bounding box annotations are available at 1Hz to 4Hz, depending on the video sequences. The labels are defined for two classes: pedestrian and car.
We built a lite-weight detection head based on YOLOX \cite{Ge2021}. Specifically, we adopt FPN \cite{Lin2017} with additional bottom-up feature fusion instead of PAFPN \cite{Ge2021}.

\vspace{0.1cm}
\noindent
\textbf{Results.}
Fig.~\ref{fig:result_objdet} compares the HMNet with the previous methods and baselines. The proposed method outperforms the state-of-the-art methods (ASTMNet \cite{Li2022} and AED \cite{Liu2022}) in both accuracy and latency.
Primarily, HMNet-B1/L1 performs remarkably well though they have only a single-level memory. For instance, HMNet-B1 performs competitively to AED \cite{Liu2022} while reducing the latency by 57\%, showing the effectiveness of the proposed ESCA. A qualitative sample in Fig.~\ref{fig:vis_all_tasks} (second row) shows the effectiveness of HMNet for detecting fast-moving objects.

Although the GRU-CSPDarknet-53 baseline achieved the best accuracy, the model has higher latency than ours (more than double compared to HMNet-L1).
More importantly, the recurrent baselines require a long accumulation time (\eg $50$ms) for constructing an event frame, which limits the frame rate of these methods.

\subsection{Monocular depth estimation}
\label{sec:depth_estimation}

\noindent
\textbf{Setups.}
Following Gehrig \etal \cite{Gehrig2021}, we pretrained our models on the synthetic Eventscape dataset \cite{Gehrig2021}. We then fine-tuned and evaluated the models on real MVSEC dataset \cite{Zhu2018MVSEC}. MVSEC dataset consists of street-scene event data and gray-scale images recorded by a DAVIS event camera with a resolution of $346\times 260$ pixels. Since the DAVIS camera is coaxial, events and gray-scale images are aligned initially.
The ground truth depth maps are recorded at 20Hz using a LiDAR sensor.
Note that they are not synchronized with the gray-scale images.
We used {\it outdoor day2} for training and {\it outdoor day1} and {\it outdoor night1} for evaluation.

\tbDepthEstimation

\vspace{0.1cm}
\noindent
\textbf{Results.}
Table~\ref{tab:depth_estimation} shows the results on MVSEC dataset.
On the daytime sequence, HMNet outperforms the previous methods while reducing latency (\eg 44\% reduction comparing HMNet-B3 and RAMNet \cite{Gehrig2021}).
Although the ResNet-50 baseline performs the best on night-time sequence, HMNet-B3 achieves competitive performance with much lower latency (65\% reduction).
On both sequences, HMNet-B3/L3 is more accurate than HMNet-B1/L1, showing the effectiveness of the multi-level memory architecture.
The qualitative results in Fig.~\ref{fig:vis_all_tasks} (bottom two rows) show that HMNet can accurately estimate the cars in the lanes.

\vspace{0.1cm}
\noindent
\textbf{Event-image fusion.}
Table~\ref{tab:depth_estimation} reports the results for event-image fusion.
The baselines show degraded accuracy with the image fusion, which is due to the temporal mismatch between the images and the ground truth.
The models need to fuse up-to-date events with an old image frame at inference timing, which is difficult to handle by naive concatenation.
On the other hand, the HMNet has dedicated architecture to handle such temporal variation and hence enjoys performance gain with the image fusion on the daytime sequence.
Meanwhile, the HMNet models perform worse with image fusion on the night-time sequence due to the image appearance gap between daytime and night-time.

\tbAblationNoiseDWCycle
\figVisEvGate
\figAblationDeltaT

\subsection{Ablation study}
\label{sec:ablation_study}
We conducted ablation studies on several key elements of HMNet using DSEC-Semantic and GEN1 datasets.

\vspace{0.1cm}
\noindent
\textbf{Effect of event gate and down-write.}
Table~\ref{tab:ablation_noise_dw_cycle} left investigates the effect of the event gate used in ESCA and the down-write operation of HMNet.
The results show the effectiveness of the components.
Applying both of them improves the accuracy by +1.6\% and +1.1\% for each dataset.
To further analyze the effect of the event gate, we visualized the attention weights of each memory cell at $\bm{z}_1$.
Specifically, Fig.~\ref{fig:vis_evgate} visualizes the accumulated attention weights written into each cell during $10$ time steps (\ie 50ms).
Without the event gate, the latent memory receives much noisy information. In contrast, with the event gate, the latent memory selectively picks up essential information about the scene (\eg the car in the green rectangle).

\vspace{0.1cm}
\noindent
\textbf{Sensitivity analysis on time step size.}
In practice, the time step size needs to be adjustable depending on the available hardware at inference time.
Hence, we performed the sensitivity analysis on the time step size.
We trained HMNet-B3/L3 using a time step size of 5ms and evaluated them with various step sizes
(3ms to 15ms).
Fig.~\ref{fig:ablation_delta_t} shows the analysis.
The models perform well until the time step size reaches 10ms
even showing better accuracy around 6ms-8ms. After that, the accuracy degrades linearly.

\vspace{0.1cm}
\noindent
\textbf{Sensitivity analysis on cycle length.}
Table~\ref{tab:ablation_noise_dw_cycle} right shows the accuracy of HMNet-B3 when the cycle length of $\bm{z}_3$ is changed between 3 to 12.
The cycle length of 9 performs better than the shorter ones, showing the importance of long-term information propagation by $\bm{z}_3$.
Though, too long cycle length impairs the accuracy on GEN1 dataset.

\section{Conclusion}
\label{sec:conclusion}
This paper proposed a novel Hierarchical Neural Memory Network (HMNet) that builds a temporal hierarchy with multi-level memories for low latency event processing. The paper also proposed an Event Sparse Cross Attention (ESCA) for embedding sparse event streams into the latent memory of HMNet with minimal information loss.
The experimental results showed that the HMNet can run faster than previous methods while achieving competitive or even better accuracy.
The experiments on event-image fusion further showed the effectiveness of HMNet for fusing temporally unaligned sensory inputs.
Our method is not limited to event data. We plan to extend our method to a variety of sensors (\eg RGB, Lider, or RGB-Lidar fusion) and broader vision applications, such as video recognition, action recognition, and pose estimation.

\section*{Acknowledgement}
This paper is based on results obtained from a project commissioned by the New Energy and Industrial Technology Development Organization (NEDO).

{\small
\bibliographystyle{ieee_fullname}
\bibliography{main}
}

\newpage
\appendix
\section{Details of the window based multi-head cross-attention (W-MCA)}
The section explains the details of W-MCA used for ``up-write'' and ``down-write'' operations.
We built W-MCA by extending the window based multi-head self-attention (W-MSA) \cite{Liu2021} with a minor modification.
We first explain W-MSA and then the modifications for our W-MCA.

\vspace{0.1cm}
\noindent
\textbf{Window based multi-head self-attention \cite{Liu2021}}.
Given an input $\bm{X}$ with a size $(H \times W \times D)$, the W-MSA computes self-attention as follows:
\begin{eqnarray}
\label{eq:w-msa}
\bm{H} = \text{W-MSA}(\bm{X})
\end{eqnarray}
Below we describe the single-head operation for simplicity since the multi-head operation can be straightforwardly acquired by applying multiple single-head operations.

In the W-MSA, query, key, and value are first calculated by Layer Normalization (LN) \cite{Ba2016} and three MLPs ($\mathcal{Q}, \mathcal{K}, \mathcal{V}$).
\begin{eqnarray}
\label{eq:norm_qkv}
\hat{\bm{X}} = \LN{\bm{X}}\quad\quad\quad\quad\quad\quad \\
\label{eq:qkv}
\bm{Q} = \mathcal{Q}(\hat{\bm{X}}),\quad
\bm{K} = \mathcal{K}(\hat{\bm{X}}),\quad
\bm{V} = \mathcal{V}(\hat{\bm{X}})
\end{eqnarray}
The query, key, and value are then divided into tiles, each with a size $(7 \times 7)$:

\begin{eqnarray}
\bm{q}_n \in \mathcal{T}_{7\times 7}(\bm{Q}),\quad
\bm{k}_n \in \mathcal{T}_{7\times 7}(\bm{K}),\quad
\bm{v}_n \in \mathcal{T}_{7\times 7}(\bm{V})
\end{eqnarray}
where $\mathcal{T}_{7\times 7}$ is a function for the tile division, and $\bm{q}_n, \bm{k}_n, \bm{v}_n$ are query, key, and value inside the $n$-th tile with a size $(49\times D)$.
Then, the attention is calculated inside each tile as follows:
\begin{eqnarray}
\bm{h}_n = {\rm softmax}(\bm{q}_n\bm{k}_n^T/\sqrt{D} + \bm{B})\bm{v}_n
\end{eqnarray}
where $\bm{B}$ is a relative position bias.
Finally, the outputs $\{\bm{h}_1, ..., \bm{h}_N\}$ from all the $N$ tiles are joined back to the original spatial size by the inverse function of the tile division:
\begin{eqnarray}
\bm{H} = \mathcal{T}^{-1}_{7\times 7}(\bm{h}_1, ..., \bm{h}_N)
\end{eqnarray}

\vspace{0.1cm}
\noindent
\textbf{Window based multi-head cross-attention}.
We build the window based multi-head cross-attention (W-MCA) to aggregate information from other memory states.
W-MCA is based on the W-MSA \cite{Liu2021} and the only difference is that we change the self-attention to the cross-attention.
Specifically, we modified Eq.~\ref{eq:w-msa}, Eq.~\ref{eq:norm_qkv}, and Eq.~\ref{eq:qkv} to have two features $\bm{X}_1, \bm{X}_2$ as inputs:
\begin{eqnarray}
\bm{H} = \text{W-MCA}(\bm{X}_1, \bm{X}_2)\quad\quad\quad \\
\hat{\bm{X}}_1 = \LN{\bm{X}_1},\quad
\hat{\bm{X}}_2 = \LN{\bm{X}_2} \quad\quad\\
\bm{Q} = \mathcal{Q}(\hat{\bm{X}_1}),\quad
\bm{K} = \mathcal{K}(\hat{\bm{X}_2}),\quad
\bm{V} = \mathcal{V}(\hat{\bm{X}_2})
\end{eqnarray}
The calculations after getting $\bm{Q}, \bm{K}, \bm{V}$ are the same as W-MSA.

\tbHyperparams
\tbHyperparamTuning
\tbConstantsDepth

\tbSemseg
\tbObjdet

\tbResultDepth

\section{Setups for semantic segmentation}
\noindent
\textbf{Dataset}.
The experiments are conducted on DSEC-Semantic dataset \cite{Sun2022}.
The dataset is a subset of DSEC dataset \cite{Gehrig2021DSEC} that consists of event camera data and RGB frames recorded at the street scene.
The resolution of the event camera and the RGB camera is $640 \times 480$ pixels and $1440 \times 1080$ pixels, respectively.
For event-image fusion, we resized the RGB images to match the resolution of event data. Note that the cameras have different viewpoints, and the RGB frames are not perfectly aligned with the event data.
The dataset contains pixel-wise annotations automatically generated from RGB images at 20Hz. In total, 8,082 and 2,809 frames are available for training and testing. Following \cite{Alonso2019}, we used 11 classes for the experiments.

\vspace{0.1cm}
\noindent
\textbf{Task head}.
We used the decoder architecture of UPerNet \cite{Xiao2018} as our task head.
For HMNet, we added bottom-up feature fusion in the task head for refreshing the high-level features with the up-to-date low-level features.
We also omitted the Pyramid Pooling Module \cite{Zhao2017} of UPerNet.

\vspace{0.1cm}
\noindent
\textbf{Training}.
Table~\ref{tab:hyperparams} shows the hyperparameters for training.
The HMNet models and the baselines are trained for 90k and 120k iterations, respectively.
We trained the recurrent baselines for $500$ iterations using Truncated Backpropagation Through Time \cite{Williams1995}, with a sequence length of 5.0sec.
We did not conduct the additional training on HMNet since it did not improve the accuracy.
We used AdamW \cite{Loshchilov2019} as the optimizer with a large weight decay coefficient of 0.01 as it performed better than Adam \cite{Kingma2015} on the task.
We used a cross-entropy loss for our loss function. We also appended auxiliary loss \cite{Zhao2017} on the output feature of $\bm{z}_3$ for HMNet and stage3 for baselines.

\section{Setups for object detection}
\noindent
\textbf{Dataset}.
The experiments are conducted on GEN1 dataset \cite{Perot2020}.
GEN1 dataset is a dataset for detecting objects from event cameras mounted on vehicles. The dataset includes 2,358 event sequences; each has a length of $60$ sec and a resolution of $304\times 240$ pixels. The sequences are divided into 1,459, 429, and 470 for training, validation, and testing.
The bounding box annotations are available at 1Hz to 4Hz, depending on the sequence. The labels are defined for two classes: pedestrian and car.

\vspace{0.1cm}
\noindent
\textbf{Task head}.
We built a lightweight detection head based on YOLOX \cite{Ge2021}. Specifically, we replaced PAFPN in YOLOX with FPN \cite{Lin2017} and added bottom-up feature fusion before top-down fusion of the FPN.

\vspace{0.1cm}
\noindent
\textbf{Training}.
On the dataset, the proposed models, the baselines, and the recurrent baselines are trained for 400k, 270k, and 135k iterations, respectively. Training more iterations did not improve the performance of the baselines.
As the labels have a low frame rate, the recurrent models require further training using a longer sequence. In this additional training, we trained HMNet and the recurrent baselines for 4.5k iterations using Truncated Backpropagation Through Time \cite{Williams1995}, with a sequence length of 8.1sec/5.0sec, respectively.

\section{Setups for depth estimation}
\noindent
\textbf{Dataset}.
Following Gehrig \etal \cite{Gehrig2021}, we pretrained our models on the synthetic Eventscape dataset \cite{Gehrig2021}. We then fine-tuned and evaluated the models on real MVSEC dataset \cite{Zhu2018MVSEC}.
Eventscape dataset consists of synthetic street-scene data generated by CARLA simulator \cite{Dosovitskiy2017}. The dataset includes event data and RGB frames with a resolution of $512\times 256$ pixels.
The ground truth depth is generated by the simulator at 25Hz, resulting in 122k, 22k, and 26k frames for training, validation, and testing.

MVSEC dataset consists of event data and gray-scale images recorded by a DAVIS event camera with a resolution of $346\times 260$ pixels, mounted on a driving car. Since the DAVIS camera is coaxial, events and gray-scale images are aligned initially.
The ground truth depth map is recorded at 20Hz using a LiDAR sensor.
The dataset includes several sequences recorded during daytime and night-time. We used {\it outdoor day2} for training and {\it outdoor day1} and {\it outdoor night1} for evaluation.
The gray-scale images are recorded at 45Hz for the daytime sequence and 10Hz for the night-time sequence.

\vspace{0.1cm}
\noindent
\textbf{Task head}.
We built the task head of the baselines based on the decoder architecture of UNet \cite{Ronneberger2015}.
Specifically, the task head applies six residual blocks on the output feature from the backbone's stage4. The task head then applies three bilinear upsampling layers, each followed by a concatenation of the skipped features from each stage and two residual blocks.
Finally, the task head applies a conv$3\times 3$ with a BatchNorm and a ReLU, conv$1\times 1$, and a sigmoid function.
The task head for HMNet has similar architecture to the baseline, but the FPN architecture replaces the UNet-like decoder part. Similar to other tasks, we added bottom-up feature fusion to the FPN.

\vspace{0.1cm}
\noindent
\textbf{Training}.
Following the previous works \cite{Carrio2020,Gehrig2021}, we trained the model to predict normalized log depth $\hat{d}$:
\begin{equation}
    \hat{d}=\frac{1}{\alpha}\log{\frac{d}{d_{\text{max}}}}+1
\end{equation}
where $d$ is a metric depth, $d_{\text{max}}$ is a maximum depth in a dataset, and $\alpha$ is a constant determined by a ratio between a maximum depth $d_{\text{max}}$ and a minimum depth $d_{\text{min}}$.
\begin{equation}
    \alpha=\log{\frac{d_{\text{max}}}{d_{\text{min}}}}
\end{equation}
Table~\ref{tab:constants_depth} shows the specific value of the constants $d_\text{max}$ and $\alpha$ for each dataset.

We trained our models with the same loss function as the previous work \cite{Gehrig2021}. Specifically, we used the scale-invariant loss \cite{Eigen2014} and the multi-scale scale-invariant gradient matching loss \cite{Li2018} for our loss function. We set a weight for the gradient matching loss as $0.25$.

\section{Details of hyperparameter tuning}
Table~\ref{tab:hyperparam_tuning} shows the search space and the result of the hyperparameter tuning on GEN1 dataset. The automatic tuning is conducted using Hyperopt \cite{Bergstra2013} with 36 iterations. Hyperopt finds a similar configuration with manual tuning.

\section{Detailed results}
Tabels~\ref{tab:semseg} and~\ref{tab:objdet} report the numerical values of the results shown in Fig.5 and Fig.6 in the main paper.
Figures~\ref{fig:supp_vis_semseg},~\ref{fig:supp_vis_objdet},~\ref{fig:supp_vis_depth_day},~\ref{fig:supp_vis_depth_night}, and~\ref{fig:supp_vis_semseg_fusion} show the additional qualitative samples.

Table~\ref{tab:result_eventscape} shows the pre-training results on the synthetic Eventscape dataset.
While HMNet and baselines perform better than previous methods on real MVSEC dataset (shown in the main paper), they perform worse in the pre-training phase.
One reason is that we apply bilinear upsampling on the model prediction instead of the convolutional upsampling used in the previous works. We find that the convolutional upsampling impairs the performance on the real MVSEC dataset while it improves accuracy on the Eventscape dataset.
Another reason might be the low temporal resolution of synthetic event data. The synthetic data has the temporal resolution of millisecond order since the data is generated based on 500Hz image frames, which might be insufficient for HMNet that works at a high operation rate (\ie 200Hz), leading to poor performance.

\figSuppVisSemseg
\figSuppVisObjdet
\figSuppVisDepthDay
\figSuppVisDepthNight
\figSuppVisSemsegFusion

\end{document}